\documentclass[11pt,a4paper]{article}
\usepackage[hyperref]{acl2021}
\usepackage{times}
\usepackage{latexsym}

\usepackage{graphicx}

\usepackage{microtype}

\aclfinalcopy 
\def\aclpaperid{2623} 

\title{DefSent: Sentence Embeddings using Definition Sentences}

\author{Hayato Tsukagoshi \hspace{6ex} Ryohei Sasano \hspace{6ex} Koichi Takeda \\
  Graduate School of Informatics, Nagoya University\\
  \texttt{tsukagoshi.hayato@e.mbox.nagoya-u.ac.jp}, \\
  \texttt{\{sasano,takedasu\}@i.nagoya-u.ac.jp} \\
  }

\date{}

\begin{document}

\maketitle

\begin{abstract}
Sentence embedding methods using natural language inference (NLI) datasets have been successfully applied to various tasks.
However, these methods are only available for limited languages due to relying heavily on the large NLI datasets.
In this paper, we propose DefSent, a sentence embedding method that uses definition sentences from a word dictionary, which performs comparably on unsupervised semantics textual similarity (STS) tasks and slightly better on SentEval tasks than conventional methods.
Since dictionaries are available for many languages, DefSent is more broadly applicable than methods using NLI datasets without constructing additional datasets.
We demonstrate that DefSent performs comparably on unsupervised semantics textual similarity (STS) tasks and slightly better on SentEval tasks to the methods using large NLI datasets.
Our code is publicly available at \url{https://github.com/hpprc/defsent}.

\end{abstract}

\section{Introduction}

Sentence embeddings represent sentences as dense vectors in a low dimensional space.
Recently, sentence embedding methods using natural language inference (NLI) datasets have been successfully applied to various tasks, including semantic textual similarity (STS) tasks. 
However, these methods are only available for limited languages due to relying heavily on the large NLI datasets.
In this paper, we propose DefSent, a sentence embedding method that uses definition sentences from a word dictionary.
Since dictionaries are available for many languages, DefSent is more broadly applicable than the methods using NLI datasets without constructing additional datasets.

\begin{figure}[t]
\centering
\includegraphics[width=7.7cm]{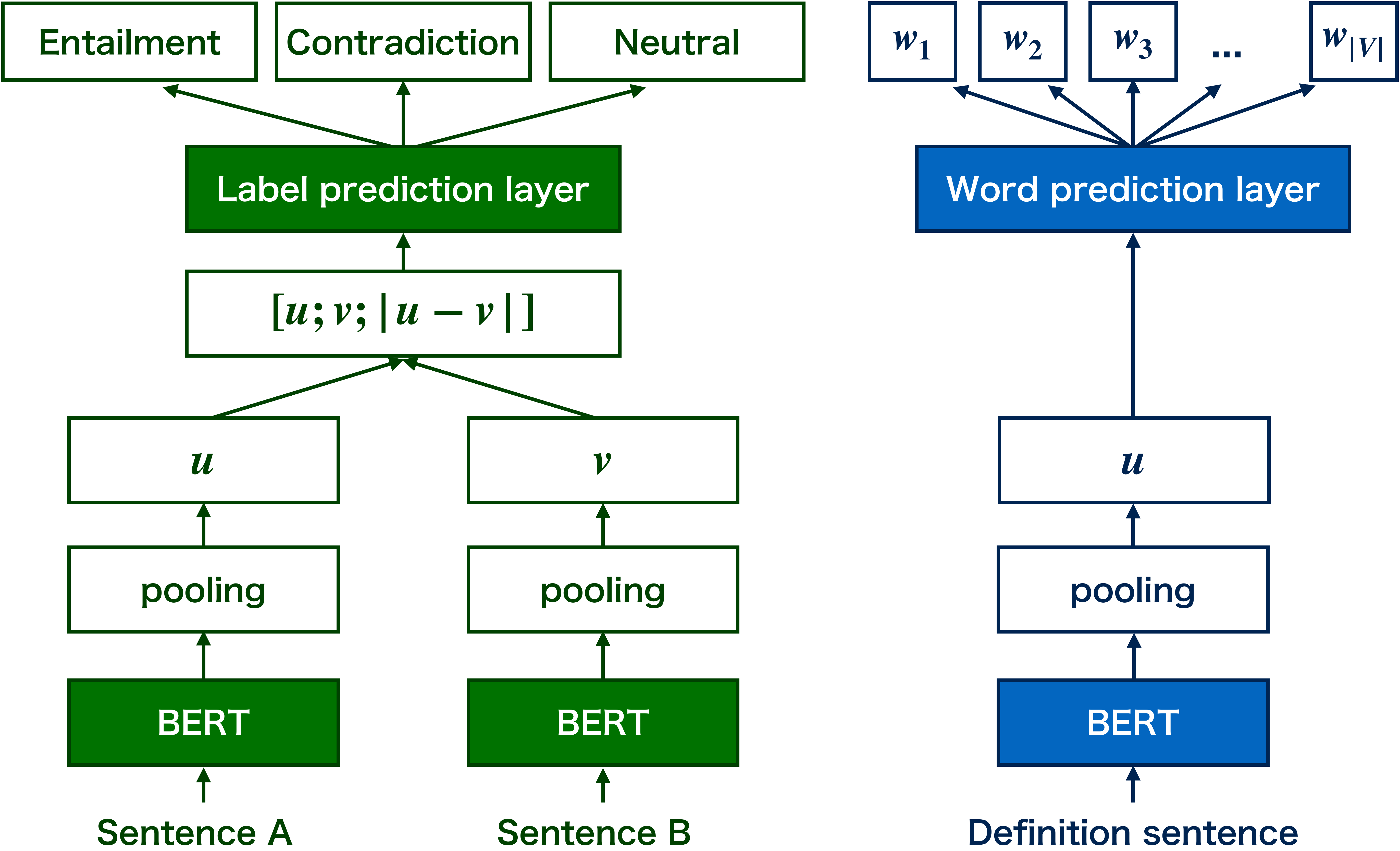}
\caption{Sentence-BERT (left) and DefSent (right).}
\vspace{-1ex}
\label{FIG::Overview}
\vspace{-1ex}
\end{figure}

Defsent is similar to the model proposed by \citet{hill-dictionary} in that it generates sentence embeddings so that the embeddings of a definition sentence and the word it represents are similar.
However, while \citet{hill-dictionary}'s model is based on recurrent neural network language models, DefSent is based on pre-trained language models such as BERT \citep{BERT} and RoBERTa \citep{RoBERTa}, with a fine-tuning mechanism as well as Sentence-BERT \citep{SBERT}.
Sentence-BERT is one of the state-of-the-art sentence embedding models, which is based on pre-trained language models that are fine-tuned on NLI datasets.
Overviews of Sentence-BERT and DefSent are depicted on Figure \ref{FIG::Overview}.

\section{Sentence Embedding Methods}
In this section, we introduce BERT, RoBERTa, and Sentence-BERT, followed by a description of DefSent, our proposed sentence embedding method.

\subsection{BERT and RoBERTa}
BERT is a pre-trained language model based on the Transformer architecture \citep{Transformer}.
Utilizing masked language modeling and next sentence prediction, BERT acquires linguistic knowledge and outputs contextualized word embeddings.
In masked language modeling, a specific proportion of input tokens is replaced with a special token \texttt{[MASK]}, and the model is trained to predict these masked tokens. 
Next sentence prediction is a task to predict whether two sentences connected by a sentence separator token \texttt{[SEP]} are consecutive sentences in the original text data.
BERT uses the output embedding of the unique token \texttt{[CLS]} at the beginning of each such sentence for prediction.

RoBERTa has the same structure as BERT. It attempts to improve BERT by removing the next sentence prediction from pre-training objectives and increasing the data size and batch size.
While both Sentence-BERT and DefSent are applicable to BERT and RoBERTa, we use BERT for the explanations in this paper.

\subsection{Sentence-BERT}

\citet{InferSent} proposed InferSent, a sentence encoder based on a Siamese network structure. InferSent trains the sentence encoder such that similar sentences are distributed close to each other in the semantic space.
\citet{SBERT} proposed Sentence-BERT, which also uses a Siamese network to create BERT-based sentence embeddings.\
An overview of Sentence-BERT is depicted on the left side of Figure \ref{FIG::Overview}.
Sentence-BERT first inputs the sentences to BERT and then constructs a sentence embedding from the output contextualized word embeddings by pooling.
They utilize the following three types of pooling strategy.

\begin{description}
  \setlength{\leftskip}{0.0cm}
  \setlength{\itemsep}{-0.05cm}
  \item[\textbf{\texttt{CLS}}] Using the \texttt{[CLS]} token embedding.; When using RoBERTa, since the \texttt{[CLS]} token does not exist, the beginning-of-sentence token \texttt{<s>} is used as an alternative.
  \item[\textbf{\texttt{Mean}}] Using the mean of the contextualized embeddings of all words in a sentence.
  \item[\textbf{\texttt{Max}}] Using the max-over-time of the contextualized embeddings of all words in a sentence.
\end{description}

Let $u$ and $v$ be the sentence embeddings for each of the sentence pairs obtained by pooling.
Then compose a vector $[u; v; |u-v|]$ and feed it to the label prediction layer, which has the same number of output dimensions as the number of classes.
For fine-tuning, \citeauthor{SBERT} uses the SNLI dataset \citep{SNLI} and the Multi-Genre NLI dataset \citep{MultiGenreNLI}, which together contain about one million sentences.

\subsection{DefSent}

Since they have the same meaning, we focus on the relationship between a definition sentence and the word it represents. To learn how to embed sentences in the semantic vector space, we train the sentence embedding model by predicting the word from definitions.
An overview of DefSent is depicted on the right side of Figure \ref{FIG::Overview}.
We call the layer that predicts the original token from the \texttt{[MASK]} embeddings used in the masked language modeling during BERT pre-training a word prediction layer.
Also, we use $w_k$ to denote the word corresponding to a given definition sentence $X_k$.

DefSent inputs the definition sentence $X_k$ to BERT and derives the sentence embedding $u$ by pooling the output embeddings.
As in Sentence-BERT, three types of pooling strategy are used: \texttt{CLS}, \texttt{Mean}, and \texttt{Max}.
Then, the derived sentence embedding $u$ is input to the word prediction layer to obtain the probability $P(w_k|X_k)$.
We use cross-entropy loss as a loss function and fine-tune BERT to maximize $P(w_k|X_k)$.

In DefSent, the parameters of the word prediction layer are fixed.
This setting allows us to fine-tune models without training an additional classifier, as is the case with both InferSent and Sentence-BERT.
Additionally, since our method uses a word prediction layer that has been pre-trained in masked language modeling, the sentence embedding $u$ is expected to be similar to the contextualized word embedding of $w_k$ when $w_k$ appears as the same meaning as $X_k$.

\section{Word Prediction Experiment}
\label{sec:d2w}

To evaluate how well DefSent can predict words from sentence embeddings, we conducted an experiment to predict a word from its definition.

\subsection{Dataset}

DefSent requires pairs of a word and its definition sentence.
We extracted these from the Oxford Dictionary dataset used by \citet{Ishiwatari}.
Each entry in the dataset consists of a word and its definition sentence, and a word can have multiple definitions. 
We split this dataset into train, dev, and test sets in the ratio of 8:1:1 word by word to evaluate how well the model can embed unseen definitions of unseen words.
It is worth noting that since DefSent utilizes the pre-trained word prediction layer of BERT and RoBERTa, it is impossible to obtain probabilities for out-of-vocabulary (OOV) words. 
Therefore, we cannot calculate losses of these OOV words in a straightforward way.\footnote{
Although we could substitute the mean of subwords as OOV word embeddings, we opted to filter out OOV words for simplicity and intuitiveness.
}
In our experiments, we only use words and their respective definitions in the dataset, as contained by the model vocabulary.
The statistics of the datasets are listed in Table \ref{tab:dataset}.

\subsection{Settings}
We used the following pre-trained models: BERT-base (bert-base-uncased), BERT-large (bert-large-uncased), RoBERTa-base (roberta-base), and RoBERTa-large (roberta-large) from Transformers \citep{Transformers}.
The batch size was 16, a fine-tuning epoch size was 1, the optimizer was Adam \citep{Adam}, and we set a linear learning rate warm-up over 10\% of the training data. 
For each respective model and pooling strategy, the learning rate was chosen based on the highest recorded Mean Reciprocal Rank (MRR) for the dev set in the range of $ 2^{x} \times 10^{-6} , x \in \{0,0.5,1,...,7\}$. 
We conducted experiments with ten different random seeds, and their mean was used as the evaluation score.
Top-$k$ accuracy (the percentage of correct answers within the first, third, and tenth positions) and MRR were calculated from the output word probabilities when a definition sentence was fed into the model.
Also, we evaluated the performance of BERT-base without fine-tuning for comparison.

\subsection{Results}

Table \ref{tab:d2w_evaluation} shows the experimental results.\footnote{
We report the fine-tuning time and computing infrastructure in Appendix \ref{appendix:runtime}, and report the learning rate, means, and standard deviations on the word prediction experiment in Appendix \ref{appendix:d2w-full}.
We also show the actual predicted words when definition sentences and other sentences are given as inputs in Appendices \ref{appendix:d2w-actual} and \ref{appendix:d2w-arbitrary}, respectively.
}
\texttt{Max} was the best pooling strategy for BERT-base without fine-tuning, but its top-$1$ accuracy was extremely low at 0.0157.
This indicates that it is not adequate for predicting words from definitions without fine-tuning.
DefSent performed higher for larger models.
In the case of BERT, \texttt{CLS} was the best pooling strategy for both base and large models.
\texttt{CLS} was also the best pooling strategy for RoBERTa-base but \texttt{Mean} was the best for RoBERTa-large.

\begin{table}[t]
\centering
\small
\tabcolsep 3pt
\begin{tabular}{l|rrr}
\hline
All & Words & Definitions & \ Avg. length  \\
\hline
\ \ Train & 29,413 & 97,759 & 9.921 \ \\
\ \ Dev & 3,677 & 12,127 & 9.874 \ \\
\ \ Test & 3,677 & 12,433 & 9.846 \ \\
\hline
\hline
In BERT vocab. & Words & Definitions & \ Avg. length  \\
\hline
\ \ Train & 7,732 & 54,142 & 9.531 \ \\
\ \ Dev & 936 & 6,544 & 9.512 \ \\
\ \ Test & 979 & 6,930 & 9.551 \ \\
\hline
\hline
In RoBERTa vocab. & Words & Definitions & \ Avg. length  \\
\hline
\ \ Train & 7,269 & 53,935 & 9.376 \ \\
\ \ Dev & 901 & 6,625 & 9.372 \ \\
\ \ Test & 925 & 6,945 & 9.410 \ \\
\hline
\end{tabular}
\vspace{-2ex}
\caption{Statistics of datasets.}\label{tab:dataset}
\vspace{-2ex}
\end{table}

\begin{table}[t]
\centering
\small
\tabcolsep 3pt
\begin{tabular}{l|l|rrrr}
\hline
Model & Pooling & MRR & Top1 & Top3 & Top10 \\
\hline
BERT-base & \texttt{CLS} & .0009 & .0000 & .0000 & .0000 \\
(no fine-tuning) & \texttt{Mean} & .0132 & .0001 & .0043 & .0242 \\
\ & \texttt{Max} & .0327 & .0157 & .0320 & .0626 \\
\hline

BERT-base   & \texttt{CLS} & .3200 & .2079 & .3670 & .5418 \\
\           & \texttt{Mean} & .3091 & .1972 & .3524 & .5356 \\
\           & \texttt{Max} & .2939 & .1840 & .3350 & .5207 \\
\hline

BERT-large  & \texttt{CLS} & \textbf{.3587} & \textbf{.2388} & \textbf{.4139} & \textbf{.6011} \\
\           & \texttt{Mean} & .3286 & .2091 & .3792 & .5723 \\
\           & \texttt{Max} & .2925 & .1814 & .3356 & .5194 \\

\hline
\hline

RoBERTa-base    & \texttt{CLS} & .3436 & .2241 & .3983 & .5836 \\
\               & \texttt{Mean} & .3365 & .2170 & .3906 & .5783 \\
\               & \texttt{Max} & .3072 & .1941 & .3523 & .5386 \\
\hline

RoBERTa-large   & \texttt{CLS} & .3863 & .2611 & .4460 & .6364 \\
\               & \texttt{Mean} & \textbf{.3995} & \textbf{.2699} & \textbf{.4634} & \textbf{.6599} \\
\               & \texttt{Max} & .3175 & .2015 & .3646 & .5543 \\
\hline
\end{tabular}
\vspace{-2ex}
\caption{Results of word prediction experiments.}
\label{tab:d2w_evaluation}
\vspace{-2ex}
\end{table}

\section{Extrinsic Evaluations}
Next, to evaluate the general quality of the constructed sentence embedding, we conducted evaluations on semantic textual similarity (STS) tasks and SentEval tasks \cite{SentEval}.

\subsection{Settings}
We compared the performance of DefSent with several existing sentence embedding methods including InferSent \cite{InferSent}, Universal Sentence Encoder \cite{USE}, and Sentence-BERT \cite{SBERT}.
For the pooling strategies, we used the strategy that achieved the highest MRR in the word prediction task for each pre-trained model.\footnote{
We report the means and standard deviations on the unsupervised STS tasks and SentEval tasks for each respective model and pooling strategy in Appendices \ref{appendix:sts-full} and \ref{appendix:senteval-full}.
}
The performance of the existing methods was taken from \citet{SBERT}.

\subsection{Semantic textual similarity tasks}

\begin{table*}[ht!]
\centering
\small
\tabcolsep 3pt
\begin{tabular}{l|c|c|c|c|c|c|c||c}
\hline
Model & STS12 & STS13 & STS14 & STS15 & STS16 & STS-B & SICK-R  & Avg.\\

\hline

Avg. GloVe embeddings \cite{GloVe} & 55.14 & 70.66 & 59.73 & 68.25 & 63.66 & 58.02 & 53.76 & 61.32\\
Avg. BERT embeddings & 38.78 & 57.98 & 57.98 & 63.15 & 61.06 & 46.35 & 58.40 & 54.81\\
BERT CLS-vector & 20.16 & 30.01 & 20.09 & 36.88 & 38.08 & 16.50 & 42.63 & 29.19\\

\hline

InferSent - Glove \cite{InferSent} & 52.86 & 66.75 & 62.15 & 72.77 & 66.87 & 68.03 & 65.65 & 65.01\\
Universal Sentence Encoder \cite{USE} & 64.49 & 67.80 & 64.61 & 76.83 & 73.18 & 74.92 & \textbf{76.69} & 71.22 \\
Sentence-BERT-base (\texttt{Mean}) & 70.97 & 76.53 & 73.19 & 79.09 & 74.30 & 77.03 & 72.91 & 74.89\\
Sentence-BERT-large (\texttt{Mean}) & 72.27 & 78.46 & \textbf{74.90} & \textbf{80.99} & 76.25 & \textbf{79.23} & 73.75 & 76.55\\
Sentence-RoBERTa-base (\texttt{Mean}) & 71.54 & 72.49 & 70.80 & 78.74 & 73.69 & 77.77 & 74.46 & 74.21\\
Sentence-RoBERTa-large (\texttt{Mean}) & \textbf{74.53} & 77.00 & 73.18 & 81.85 & 76.82 & 79.10 & 74.29 & \textbf{76.68}\\

\hline

DefSent-BERT-base (\texttt{CLS}) & 67.56 & 79.86 & 69.52 & 76.83 & 76.61 & 75.57 & 73.05 & 74.14 \\
DefSent-BERT-large (\texttt{CLS}) & 66.22 & \textbf{82.07} & 71.48 & 79.34 & 75.38 & 73.46 & 74.30 & 74.61 \\
DefSent-RoBERTa-base (\texttt{CLS}) & 65.55 & 80.84 & 71.87 & 78.77 & \textbf{79.29} & 78.13 & 74.92 & 75.62 \\
DefSent-RoBERTa-large (\texttt{Mean}) & 58.36 & 76.24 & 69.55 & 73.15 & 76.90 & 78.53 & 73.81 & 72.36 \\

\hline
\end{tabular}
\vspace{-2ex}
\caption{
Spearman's rank correlation $\rho \times 100$ between cosine similarities of sentence embeddings and human ratings. \
STS-B denotes STS Benchmark, and SICK-R denotes SICK-Relatedness.
}
\vspace{1ex}
\label{table:sts_evaluation}
\end{table*}

\begin{table*}[ht!]
\centering
\small
\tabcolsep 5pt
\begin{tabular}{l|c|c|c|c|c|c|c||c}

\hline
Model & MR & CR & SUBJ & MPQA & SST-2 & TREC & MRPC & Avg. \\

\hline
Avg. GloVe embeddings & 77.25 & 78.30 & 91.17 & 87.85 & 80.18 & 83.00 & 72.87 & 81.52 \\
Avg. BERT embeddings & 78.66 & 86.25 & 94.37 & 88.66 & 84.40 & 92.80 & 69.45 & 84.94 \\
BERT CLS-vector & 78.68 & 84.85 & 94.21 & 88.23 & 84.13 & 91.40 & 71.13 & 84.66 \\

\hline
InferSent - GloVe & 81.57 & 86.54 & 92.50 & 90.38 & 84.18 & 88.20 & 75.77 & 85.59 \\
Universal Sentence Encoder & 80.09 & 85.19 & 93.98 & 86.70 & 86.38 & \textbf{93.20} & 70.14 & 85.10 \\
Sentence-BERT-base (\texttt{Mean}) & 83.64 & 89.43 & 94.39 & 89.86 & 88.96 & 89.60 & \textbf{76.00} & 87.41 \\
Sentence-BERT-large (\texttt{Mean}) & 84.88 & 90.07 & 94.52 & 90.33 & 90.66 & 87.40 & 75.94 & 87.69 \\

\hline
DefSent-BERT-base (\texttt{CLS}) & 80.94 & 87.57 & 94.59 & 89.98 & 85.78 & 89.73 & 73.82 & 86.06 \\
DefSent-BERT-large (\texttt{CLS}) & 85.79 & 90.54 & \textbf{95.58} & 90.15 & \textbf{91.17} & 90.47 & 73.74 & 88.20 \\
DefSent-RoBERTa-base (\texttt{CLS}) & 83.94 & 90.44 & 94.05 & 90.70 & 89.16 & 90.80 & 75.52 & 87.80 \\
DefSent-RoBERTa-large (\texttt{Mean}) & \textbf{86.47} & \textbf{91.53} & 95.02 & \textbf{91.15} & 90.77 & 92.33 & 73.91 & \textbf{88.74} \\
\hline
\end{tabular}
\vspace{-2ex}
\caption{
Accuracy (\%) for each task in SentEval.
}
\vspace{1ex}
\label{table:senteval}
\end{table*}

We evaluated DefSent on unsupervised STS tasks.
In these tasks, we compute semantic similarities of given sentence pairs and calculate Spearman's rank correlation $\rho$ between similarities and gold scores of sentence similarities.
In the unsupervised setting, none of the models are optimized on the STS datasets.
Instead, the similarities of the given sentence embeddings are calculated using common similarity measures such as negative Manhattan distance, negative Euclidean distance, and cosine-similarity. 
In this study, we used cosine-similarity.

We performed experiments on unsupervised STS tasks using the STS12-16 \citep{STS12, STS13, STS14, STS15, STS16}, STS Benchmark \citep{STSB}, and SICK-Relatedness \citep{SICK} datasets.
These datasets contain sentence pairs and their similarity scores, which is a real number from 0 to 5 assigned by human evaluations.
Experiments were conducted with ten different random seeds, and the mean was used as the evaluation score.

Table \ref{table:sts_evaluation} shows the experimental results.
Although the training data size used in DefSent was only about 5\% that of Sentence-BERT, DefSent-BERT-base and DefSent-RoBERTa-base performed comparably to Sentence-BERT-base and Sentence-RoBERTa-base.
In particular, DefSent-RoBERTa models showed high performance in the STS Benchmark.

\subsection{SentEval}

SentEval \cite{SentEval} is a popular toolkit for evaluating the quality of universal sentence embeddings that aggregates various tasks, including binary and multi-class classification, natural language inference, and sentence similarity.
For the SentEval evaluations, we trained a logistic regression classifier using sentence embeddings as input features to evaluate the extent to which each sentence embedding contained the important information for each task. 
We used the same tasks and settings as \citet{SBERT} and performed a 10-fold cross-validation. 
We conducted experiments with three different random seeds, and the mean was used as the evaluation score.

Table \ref{table:senteval} shows the results.\footnote{
\citet{SBERT} reported that there were minor difference from Sentence-BERT, so we omitted the results of Sentence-RoBERTa.
}
DefSent-RoBERTa-large achieved the best average score among all models. 
Also, increasing the model size improved the performance consistently.
The performances of DefSent-BERT-large, DefSent-RoBERTa-base, and DefSent-RoBERTa-large were better than the performances of Sentence-BERT-based methods.
These results indicate that DefSent embeds useful information that can be applied to various tasks.

\section{Conclusion}
In this paper, we proposed DefSent, a new sentence embedding method using a dictionary, and demonstrated its effectiveness through a series of experiments.
Its performance was comparable to or even slightly better than existing methods using large NLI datasets.
DefSent is based on dictionaries developed for many languages, so it does not require new language resources when applied to other languages.
Since the model is trained with the same word prediction process as the masked language modeling, sentence embeddings derived by DefSent are expected to be similar to contextualized word embeddings of a word when it appears with the same meaning as the definition. 

In future work, we will evaluate the performance of DefSent when it is applied to languages other than English and when it is applied to a broader range of downstream tasks, such as document classification tasks. 
We will also analyze the relationship between the sentence embeddings by DefSent and the contextualized word embeddings in the semantic vector space and investigate how model architecture and size influence the embeddings.

\section*{Acknowledgements}
This work was supported by JSPS KAKENHI Grant Number 21H04901.

\bibliographystyle{acl_natbib}
\bibliography{acl2021}

\begin{thebibliography}{21}
\expandafter\ifx\csname natexlab\endcsname\relax\def\natexlab#1{#1}\fi

\bibitem[{Agirre et~al.(2015)Agirre, Banea, Cardie, Cer, Diab, Gonzalez-Agirre,
  Guo, Lopez-Gazpio, Maritxalar, Mihalcea, Rigau, Uria, and Wiebe}]{STS15}
Eneko Agirre, Carmen Banea, Claire Cardie, Daniel Cer, Mona Diab, Aitor
  Gonzalez-Agirre, Weiwei Guo, I{\~n}igo Lopez-Gazpio, Montse Maritxalar, Rada
  Mihalcea, German Rigau, Larraitz Uria, and Janyce Wiebe. 2015.
\newblock \href {https://doi.org/10.18653/v1/S15-2045} {{S}em{E}val-2015 {T}ask
  2: {S}emantic {T}extual {S}imilarity, {E}nglish, {S}panish and {P}ilot on
  {I}nterpretability}.
\newblock In \emph{Proceedings of the 9th International Workshop on Semantic
  Evaluation ({S}em{E}val)}, pages 252--263.

\bibitem[{Agirre et~al.(2014)Agirre, Banea, Cardie, Cer, Diab, Gonzalez-Agirre,
  Guo, Mihalcea, Rigau, and Wiebe}]{STS14}
Eneko Agirre, Carmen Banea, Claire Cardie, Daniel Cer, Mona Diab, Aitor
  Gonzalez-Agirre, Weiwei Guo, Rada Mihalcea, German Rigau, and Janyce Wiebe.
  2014.
\newblock \href {https://doi.org/10.3115/v1/S14-2010} {{S}em{E}val-2014 {T}ask
  10: {M}ultilingual {S}emantic {T}extual {S}imilarity}.
\newblock In \emph{Proceedings of the 8th International Workshop on Semantic
  Evaluation ({S}em{E}val)}, pages 81--91.

\bibitem[{Agirre et~al.(2016)Agirre, Banea, Cer, Diab, Gonzalez-Agirre,
  Mihalcea, Rigau, and Wiebe}]{STS16}
Eneko Agirre, Carmen Banea, Daniel Cer, Mona Diab, Aitor Gonzalez-Agirre, Rada
  Mihalcea, German Rigau, and Janyce Wiebe. 2016.
\newblock \href {https://doi.org/10.18653/v1/S16-1081} {{S}em{E}val-2016 {T}ask
  1: {S}emantic {T}extual {S}imilarity, {M}onolingual and {C}ross-{L}ingual
  {E}valuation}.
\newblock In \emph{Proceedings of the 10th International Workshop on Semantic
  Evaluation ({S}em{E}val)}, pages 497--511.

\bibitem[{Agirre et~al.(2012)Agirre, Cer, Diab, and Gonzalez-Agirre}]{STS12}
Eneko Agirre, Daniel Cer, Mona Diab, and Aitor Gonzalez-Agirre. 2012.
\newblock \href {https://www.aclweb.org/anthology/S12-1051} {{S}em{E}val-2012
  {T}ask 6: {A} {P}ilot on {S}emantic {T}extual {S}imilarity}.
\newblock In \emph{*{SEM} 2012: The First Joint Conference on Lexical and
  Computational Semantics {--} Semantic Evaluation ({S}em{E}val)}, pages
  385--393.

\bibitem[{Agirre et~al.(2013)Agirre, Cer, Diab, Gonzalez-Agirre, and
  Guo}]{STS13}
Eneko Agirre, Daniel Cer, Mona Diab, Aitor Gonzalez-Agirre, and Weiwei Guo.
  2013.
\newblock \href {https://www.aclweb.org/anthology/S13-1004} {*{SEM} 2013 shared
  task: {S}emantic {T}extual {S}imilarity}.
\newblock In \emph{Second Joint Conference on Lexical and Computational
  Semantics (*{SEM})}, pages 32--43.

\bibitem[{Bowman et~al.(2015)Bowman, Angeli, Potts, and Manning}]{SNLI}
Samuel~R. Bowman, Gabor Angeli, Christopher Potts, and Christopher~D. Manning.
  2015.
\newblock \href {https://doi.org/10.18653/v1/D15-1075} {A large annotated
  corpus for learning natural language inference}.
\newblock In \emph{Proceedings of the 2015 Conference on Empirical Methods in
  Natural Language Processing (EMNLP)}, pages 632--642.

\bibitem[{Cer et~al.(2017)Cer, Diab, Agirre, Lopez-Gazpio, and Specia}]{STSB}
Daniel Cer, Mona Diab, Eneko Agirre, I{\~n}igo Lopez-Gazpio, and Lucia Specia.
  2017.
\newblock \href {https://doi.org/10.18653/v1/S17-2001} {{S}em{E}val-2017 {T}ask
  1: {S}emantic {T}extual {S}imilarity {M}ultilingual and {C}rosslingual
  {F}ocused {E}valuation}.
\newblock In \emph{Proceedings of the 11th International Workshop on Semantic
  Evaluation ({S}em{E}val)}, pages 1--14.

\bibitem[{Cer et~al.(2018)Cer, Yang, yi~Kong, Hua, Limtiaco, John, Constant,
  Guajardo-Cespedes, Yuan, Tar, Sung, Strope, and Kurzweil}]{USE}
Daniel~Matthew Cer, Yinfei Yang, Sheng yi~Kong, Nan Hua, Nicole Limtiaco,
  Rhomni~St. John, Noah Constant, Mario Guajardo-Cespedes, Steve Yuan, C.~Tar,
  Yun-Hsuan Sung, B.~Strope, and R.~Kurzweil. 2018.
\newblock \href {http://arxiv.org/abs/1803.11175} {{U}niversal {S}entence
  {E}ncoder}.
\newblock \emph{arXiv:1803.11175}.

\bibitem[{Conneau and Kiela(2018)}]{SentEval}
Alexis Conneau and Douwe Kiela. 2018.
\newblock \href {https://www.aclweb.org/anthology/L18-1269} {{S}ent{E}val: {A}n
  {E}valuation {T}oolkit for {U}niversal {S}entence {R}epresentations}.
\newblock In \emph{Proceedings of the Eleventh International Conference on
  Language Resources and Evaluation ({LREC})}, pages 1699--1704.

\bibitem[{Conneau et~al.(2017)Conneau, Kiela, Schwenk, Barrault, and
  Bordes}]{InferSent}
Alexis Conneau, Douwe Kiela, Holger Schwenk, Lo{\"\i}c Barrault, and Antoine
  Bordes. 2017.
\newblock \href {https://doi.org/10.18653/v1/D17-1070} {{S}upervised {L}earning
  of {U}niversal {S}entence {R}epresentations from {N}atural {L}anguage
  {I}nference {D}ata}.
\newblock In \emph{Proceedings of the 2017 Conference on Empirical Methods in
  Natural Language Processing (EMNLP)}, pages 670--680.

\bibitem[{Devlin et~al.(2019)Devlin, Chang, Lee, and Toutanova}]{BERT}
Jacob Devlin, Ming-Wei Chang, Kenton Lee, and Kristina Toutanova. 2019.
\newblock \href {https://doi.org/10.18653/v1/N19-1423} {{BERT}: {P}re-training
  of {D}eep {B}idirectional {T}ransformers for {L}anguage {U}nderstanding}.
\newblock In \emph{Proceedings of the 2019 Conference of the North {A}merican
  Chapter of the Association for Computational Linguistics: Human Language
  Technologies (NAACL)}, pages 4171--4186.

\bibitem[{Hill et~al.(2016)Hill, Cho, Korhonen, and Bengio}]{hill-dictionary}
Felix Hill, Kyunghyun Cho, Anna Korhonen, and Yoshua Bengio. 2016.
\newblock \href {https://doi.org/10.1162/tacl_a_00080} {{L}earning to
  {U}nderstand {P}hrases by {E}mbedding the {D}ictionary}.
\newblock In \emph{Transactions of the Association for Computational
  Linguistics (TACL)}, pages 17--30.

\bibitem[{Ishiwatari et~al.(2019)Ishiwatari, Hayashi, Yoshinaga, Neubig, Sato,
  Toyoda, and Kitsuregawa}]{Ishiwatari}
Shonosuke Ishiwatari, Hiroaki Hayashi, Naoki Yoshinaga, Graham Neubig, Shoetsu
  Sato, Masashi Toyoda, and Masaru Kitsuregawa. 2019.
\newblock \href {https://doi.org/10.18653/v1/N19-1350} {Learning to {D}escribe
  {U}nknown {P}hrases with {L}ocal and {G}lobal {C}ontexts}.
\newblock In \emph{Proceedings of the 2019 Conference of the North {A}merican
  Chapter of the Association for Computational Linguistics: Human Language
  Technologies ({NAACL})}, pages 3467--3476.

\bibitem[{Kingma and Ba(2015)}]{Adam}
Diederik~P. Kingma and Jimmy Ba. 2015.
\newblock \href {http://arxiv.org/abs/1412.6980} {{A}dam: {A} {M}ethod for
  {S}tochastic {O}ptimization}.
\newblock In \emph{3rd International Conference on Learning Representations
  ({ICLR})}.

\bibitem[{Liu et~al.(2019)Liu, Ott, Goyal, Du, Joshi, Chen, Levy, Lewis,
  Zettlemoyer, and Stoyanov}]{RoBERTa}
Yinhan Liu, Myle Ott, Naman Goyal, Jingfei Du, Mandar Joshi, Danqi Chen, Omer
  Levy, M.~Lewis, Luke Zettlemoyer, and Veselin Stoyanov. 2019.
\newblock \href {http://arxiv.org/abs/1907.11692} {{R}o{BERT}a: {A} {R}obustly
  {O}ptimized {BERT} {P}retraining {A}pproach}.
\newblock \emph{arXiv:1907.11692}.

\bibitem[{Marelli et~al.(2014)Marelli, Menini, Baroni, Bentivogli, Bernardi,
  and Zamparelli}]{SICK}
Marco Marelli, Stefano Menini, Marco Baroni, Luisa Bentivogli, Raffaella
  Bernardi, and Roberto Zamparelli. 2014.
\newblock \href
  {http://www.lrec-conf.org/proceedings/lrec2014/pdf/363_Paper.pdf} {A {SICK}
  cure for the evaluation of compositional distributional semantic models}.
\newblock In \emph{Proceedings of the Ninth International Conference on
  Language Resources and Evaluation ({LREC})}, pages 216--223.

\bibitem[{Pennington et~al.(2014)Pennington, Socher, and Manning}]{GloVe}
Jeffrey Pennington, Richard Socher, and Christopher Manning. 2014.
\newblock \href {https://doi.org/10.3115/v1/D14-1162} {{G}lo{V}e: {G}lobal
  {V}ectors for {W}ord {R}epresentation}.
\newblock In \emph{Proceedings of the 2014 Conference on Empirical Methods in
  Natural Language Processing ({EMNLP})}, pages 1532--1543.

\bibitem[{Reimers and Gurevych(2019)}]{SBERT}
Nils Reimers and Iryna Gurevych. 2019.
\newblock \href {https://doi.org/10.18653/v1/D19-1410} {Sentence-{BERT}:
  {S}entence {E}mbeddings using {S}iamese {BERT}-{N}etworks}.
\newblock In \emph{Proceedings of the 2019 Conference on Empirical Methods in
  Natural Language Processing and the 9th International Joint Conference on
  Natural Language Processing (EMNLP-IJCNLP)}, pages 3982--3992.

\bibitem[{Vaswani et~al.(2017)Vaswani, Shazeer, Parmar, Uszkoreit, Jones,
  Gomez, Kaiser, and Polosukhin}]{Transformer}
Ashish Vaswani, Noam Shazeer, Niki Parmar, Jakob Uszkoreit, Llion Jones,
  Aidan~N Gomez, {\L}ukasz Kaiser, and Illia Polosukhin. 2017.
\newblock \href
  {https://proceedings.neurips.cc/paper/2017/file/3f5ee243547dee91fbd053c1c4a845aa-Paper.pdf}
  {{A}ttention is {A}ll you {N}eed}.
\newblock In \emph{Advances in Neural Information Processing Systems (NIPS)},
  pages 5998--6008.

\bibitem[{Williams et~al.(2018)Williams, Nangia, and Bowman}]{MultiGenreNLI}
Adina Williams, Nikita Nangia, and Samuel Bowman. 2018.
\newblock \href {https://doi.org/10.18653/v1/N18-1101} {A {B}road-{C}overage
  {C}hallenge {C}orpus for {S}entence {U}nderstanding through {I}nference}.
\newblock In \emph{Proceedings of the 2018 Conference of the North {A}merican
  Chapter of the Association for Computational Linguistics: Human Language
  Technologies (NAACL)}, pages 1112--1122.

\bibitem[{Wolf et~al.(2020)Wolf, Debut, Sanh, Chaumond, Delangue, Moi, Cistac,
  Rault, Louf, Funtowicz, Davison, Shleifer, von Platen, Ma, Jernite, Plu, Xu,
  Scao, Gugger, Drame, Lhoest, and Rush}]{Transformers}
Thomas Wolf, Lysandre Debut, Victor Sanh, Julien Chaumond, Clement Delangue,
  Anthony Moi, Pierric Cistac, Tim Rault, Rémi Louf, Morgan Funtowicz, Joe
  Davison, Sam Shleifer, Patrick von Platen, Clara Ma, Yacine Jernite, Julien
  Plu, Canwen Xu, Teven~Le Scao, Sylvain Gugger, Mariama Drame, Quentin Lhoest,
  and Alexander~M. Rush. 2020.
\newblock \href {https://www.aclweb.org/anthology/2020.emnlp-demos.6}
  {{T}ransformers: {S}tate-of-the-{A}rt {N}atural {L}anguage {P}rocessing}.
\newblock In \emph{Proceedings of the 2020 Conference on Empirical Methods in
  Natural Language Processing (EMNLP): System Demonstrations}, pages 38--45.

\end{thebibliography}


\appendix

\newcommand{\pmstd}[1]{{\tiny $\pm{#1}$}}

\begin{table*}[ht!]
\centering
\small
\tabcolsep 2.8pt

\begin{tabular}{@{ }l@{ }|l|c|c|c|c|c}
\hline

Model & Pooling & Learning rate & MRR & Top1 & Top3 & Top10 \\

\hline

\rule{0pt}{2.2ex} BERT-base & \texttt{CLS} & $2^{2.5}\times 10^{-6}$ & .3200\pmstd{.0020} & .2079\pmstd{.0021} & .3670\pmstd{.0029} & .5418\pmstd{.0022} \\
\ & \texttt{Mean} & $2^{3.5}\times 10^{-6}$ & .3091\pmstd{.0021} & .1972\pmstd{.0030} & .3524\pmstd{.0038} & .5356\pmstd{.0029} \\
\ & \texttt{Max} & $2^{3.5}\times 10^{-6}$ & .2939\pmstd{.0021} & .1840\pmstd{.0026} & .3350\pmstd{.0023} & .5207\pmstd{.0045} \\

\hline

\rule{0pt}{2.2ex} BERT-large & \texttt{CLS} & $2^{2.5}\times 10^{-6}$ & .3587\pmstd{.0043} & .2388\pmstd{.0047} & .4139\pmstd{.0059} & .6011\pmstd{.0054} \\
\ & \texttt{Mean} & $2^{3.5}\times 10^{-6}$ & .3286\pmstd{.0044} & .2091\pmstd{.0045} & .3792\pmstd{.0055} & .5723\pmstd{.0072} \\
\ & \texttt{Max} & $2^{3.0}\times 10^{-6}$ & .2925\pmstd{.0138} & .1814\pmstd{.0113} & .3356\pmstd{.0172} & .5194\pmstd{.0181} \\

\hline
\hline

\rule{0pt}{2.2ex} RoBERTa-base & \texttt{CLS} & $2^{2.5}\times 10^{-6}$ & .3436\pmstd{.0016} & .2241\pmstd{.0016} & .3983\pmstd{.0027} & .5836\pmstd{.0017} \\
\ & \texttt{Mean} & $2^{3.0}\times 10^{-6}$ & .3365\pmstd{.0017} & .2170\pmstd{.0014} & .3906\pmstd{.0029} & .5783\pmstd{.0022} \\
\ & \texttt{Max} & $2^{2.0}\times 10^{-6}$ & .3072\pmstd{.0037} & .1941\pmstd{.0039} & .3523\pmstd{.0050} & .5386\pmstd{.0064} \\

\hline

\rule{0pt}{2.2ex} RoBERTa-large & \texttt{CLS} & $2^{2.0}\times 10^{-6}$ & .3863\pmstd{.0040} & .2611\pmstd{.0045} & .4460\pmstd{.0044} & .6364\pmstd{.0041} \\
\ & \texttt{Mean} & $2^{2.0}\times 10^{-6}$ & .3995\pmstd{.0041} & .2699\pmstd{.0053} & .4634\pmstd{.0042} & .6599\pmstd{.0036} \\
\ & \texttt{Max} & $2^{2.5}\times 10^{-6}$ & .3175\pmstd{.0069} & .2015\pmstd{.0054} & .3646\pmstd{.0087} & .5543\pmstd{.0092} \\

\hline

\end{tabular}
\caption{
MRR, top-$1$, top-$3$, and top-$10$ accuracy on the word prediction experiment. The scores are the mean and standard deviation of 10 evaluations with different random seeds.
}
\label{table:word_prediction_full}
\end{table*}

\begin{table*}[t]
\centering
\small
\begin{tabular}{l||l||l|l|l}
\hline
Word & Definition & \multicolumn{3}{l}{Predictions (1st, 2nd, 3rd)} \\
\hline
\hline
cost & be expensive for ( someone ) & \textbf{cost} & charge & pay \\
\hline
preserve & prevent ( food ) from rotting & \textbf{preserve} & keep & spoil \\
\hline
good & that which is pleasing or valuable or useful & \textbf{good} & pleasing & pleasure \\
\hline
linux & an open-source operating system modelled on unix. & \textbf{linux} & unix & gnu \\
\hline
pile & place or lay as if in a pile & \textbf{pile} & stack & heap \\
\hline
weird & very strange; bizarre & \textbf{weird} & strange & bizarre \\
\hline
sale & the general activity of selling & selling & \textbf{sale} & retail \\
\hline
satellite & a celestial body orbiting the earth or another planet. & planet & \textbf{satellite} & orbit \\
\hline
\hline
logic & the quality of being justifiable by reason & reason & justice & certainty \\
\hline
custom & a thing that one does habitually & habit & routine & ritual \\
\hline
chief & a person who is in charge & leader & boss & master \\
\hline
nirvana & an ideal or idyllic state or place & paradise & dream & ideal \\
\hline
\end{tabular}
\caption{Predicted words when the embeddings of definition sentences are input. 
The first two columns represent words and their defining sentences, and the third to fifth columns represent the top three predicted words. Correctly predicted words shown in \textbf{bold}.
}
\label{tab:d2w_actual_results}
\end{table*}

\begin{table*}[t]
\centering
\small
\begin{tabular}{l||l|l|l|l|l}
\hline
Input & \multicolumn{5}{l}{Predictions (1st, 2nd, 3rd, 4th, 5th)} \\
\hline
\hline
royal man & king & royal & prince & noble & knight\\
\hline
royal woman & queen & princess & royal  & regal & sovereign\\
\hline
royal boy & boy & prince & royal & king  & baby\\
\hline
royal girl & princess & queen & lady & royal & belle\\
\hline
\hline
good & fine & good & great & right & solid\\
\hline
bad & bad & dirty & awful & ugly & nasty\\
\hline
not good & bad & poor & wrong & awful & terrible\\
\hline
not bad & okay & fair & good & fine & ok\\
\hline
\hline
Star wars & jedi & star & trek & galaxy & saga\\
\hline
Star wars in America & jedi & western & fan & hollywood & movie\\
\hline
Star wars in Europe & trek & space & adventure & cinema & fantas\\
\hline
Star wars in Japan & godzilla & anime & gundam & jedi & manga\\
\hline
captain america & marvel & hero & thor & superhero & hulk\\
\hline
\end{tabular}
\caption{Predicted words when the embeddings of sentences other than definition sentences are input.
}
\label{tab:d2w_arbitrary_results}
\end{table*}

\begin{table*}[ht!]
\centering
\small
\tabcolsep 2.5pt
\begin{tabular}{@{ }l@{ }|l|c|c|c|c|c|c|c||c}
\hline

Model & Pooling & STS12 & STS13 & STS14 & STS15 & STS16 & STS-B & SICK-R  & Avg.\\

\hline

\rule{0pt}{2.0ex}BERT-base & \texttt{CLS} & 67.56\pmstd{0.26} & 79.86\pmstd{0.25} & 69.52\pmstd{0.39} & 76.83\pmstd{0.32} & 76.61\pmstd{0.33} & 75.57\pmstd{0.37} & 73.05\pmstd{0.32} & 74.14\pmstd{0.25} \\
\ & \texttt{Mean} & 67.30\pmstd{0.44} & 81.96\pmstd{0.24} & 71.92\pmstd{0.28} & 77.68\pmstd{0.47} & 76.71\pmstd{0.48} & 76.90\pmstd{0.40} & 73.28\pmstd{0.30} & 75.11\pmstd{0.21} \\
\ & \texttt{Max} & 64.61\pmstd{0.87} & 82.06\pmstd{0.21} & 72.43\pmstd{0.31} & 76.56\pmstd{0.74} & 75.61\pmstd{0.43} & 76.61\pmstd{0.52} & 72.15\pmstd{0.46} & 74.29\pmstd{0.33} \\

\hline

\rule{0pt}{2.0ex}BERT-large & \texttt{CLS} & 66.22\pmstd{0.79} & 82.07\pmstd{0.39} & 71.48\pmstd{0.33} & 79.34\pmstd{0.44} & 75.38\pmstd{0.60} & 73.46\pmstd{0.45} & 74.30\pmstd{0.50} & 74.61\pmstd{0.41} \\
\ & \texttt{Mean} & 64.18\pmstd{0.96} & 82.76\pmstd{0.42} & 73.14\pmstd{0.32} & 79.66\pmstd{0.92} & 77.93\pmstd{0.78} & 77.89\pmstd{0.89} & 73.98\pmstd{0.46} & 75.65\pmstd{0.53} \\
\ & \texttt{Max} & 58.94\pmstd{1.06} & 81.03\pmstd{0.66} & 71.34\pmstd{0.88} & 76.23\pmstd{1.83} & 76.07\pmstd{0.56} & 75.75\pmstd{0.70} & 71.69\pmstd{0.74} & 73.01\pmstd{0.74} \\

\hline

\rule{0pt}{2.0ex}RoBERTa-base & \texttt{CLS} & 65.55\pmstd{0.89} & 80.84\pmstd{0.26} & 71.87\pmstd{0.39} & 78.77\pmstd{0.70} & 79.29\pmstd{0.27} & 78.13\pmstd{0.61} & 74.92\pmstd{0.18} & 75.62\pmstd{0.38} \\
\ & \texttt{Mean} & 60.78\pmstd{1.41} & 77.17\pmstd{0.60} & 69.71\pmstd{0.73} & 75.13\pmstd{1.00} & 77.75\pmstd{0.38} & 76.52\pmstd{0.63} & 74.10\pmstd{0.45} & 73.02\pmstd{0.63} \\
\ & \texttt{Max} & 63.85\pmstd{0.86} & 78.55\pmstd{0.90} & 71.19\pmstd{0.86} & 76.55\pmstd{1.12} & 77.86\pmstd{0.59} & 78.02\pmstd{0.77} & 73.97\pmstd{0.46} & 74.28\pmstd{0.62} \\

\hline

\rule{0pt}{2.0ex}RoBERTa-large & \texttt{CLS} & 63.84\pmstd{1.34} & 77.33\pmstd{2.53} & 68.64\pmstd{1.34} & 72.86\pmstd{1.96} & 77.13\pmstd{1.32} & 78.32\pmstd{1.08} & 74.14\pmstd{1.31} & 73.18\pmstd{1.20} \\
\ & \texttt{Mean} & 58.36\pmstd{1.16} & 76.24\pmstd{0.87} & 69.55\pmstd{0.85} & 73.15\pmstd{1.32} & 76.90\pmstd{0.94} & 78.53\pmstd{0.54} & 73.81\pmstd{0.88} & 72.36\pmstd{0.73} \\
\ & \texttt{Max} & 62.89\pmstd{1.42} & 77.99\pmstd{1.88} & 69.83\pmstd{1.66} & 75.60\pmstd{1.51} & 79.63\pmstd{0.60} & 79.34\pmstd{0.48} & 74.04\pmstd{0.84} & 74.19\pmstd{0.88} \\

\hline

\end{tabular}
\caption{
Spearman's rank correlation $\rho \times 100$ between the cosine similarities of the sentence embeddings and the human ratings for each model and pooling strategy.
The scores are the mean and standard deviation of 10 evaluations with different random seeds.
}
\label{table:sts_full}
\end{table*}

\begin{table*}[ht!]
\centering
\small
\tabcolsep 2.5pt
\begin{tabular}{@{ }l@{ }|l|c|c|c|c|c|c|c||c}
\hline

Model & Pooling & MR & CR & SUBJ & MPQA & SST-2 & TREC & MRPC & Avg. \\

\hline

\rule{0pt}{2.0ex}BERT-base & \texttt{CLS} & 80.94\pmstd{0.08} & 87.57\pmstd{0.12} & 94.59\pmstd{0.09} & 89.98\pmstd{0.04} & 85.78\pmstd{1.14} & 89.73\pmstd{0.76} & 73.82\pmstd{0.19} & 86.06\pmstd{0.28} \\
\ & \texttt{Mean} &81.84\pmstd{0.17} & 88.20\pmstd{0.04} & 94.82\pmstd{0.12} & 89.94\pmstd{0.12} & 86.49\pmstd{0.20} & 89.73\pmstd{0.31} & 75.32\pmstd{0.78} & 86.62\pmstd{0.18}\\
\ & \texttt{Max} & 80.74\pmstd{0.16} & 88.00\pmstd{0.09} & 94.32\pmstd{0.07} & 89.92\pmstd{0.25} & 85.03\pmstd{0.09} & 89.13\pmstd{0.50} & 74.11\pmstd{0.49} & 85.89\pmstd{0.02}\\

\hline
\rule{0pt}{2.0ex}BERT-large & \texttt{CLS} & 85.79\pmstd{0.19} & 90.54\pmstd{0.26} & 95.58\pmstd{0.14} & 90.15\pmstd{0.04} & 91.17\pmstd{0.06} & 90.47\pmstd{0.95} & 73.74\pmstd{0.61} & 88.20\pmstd{0.07}\\
\ & \texttt{Mean} & 84.05\pmstd{0.25} & 89.50\pmstd{0.24} & 95.21\pmstd{0.12} & 90.19\pmstd{0.36} & 89.44\pmstd{0.14} & 88.60\pmstd{0.87} & 73.99\pmstd{0.90} & 87.28\pmstd{0.05}\\
\ & \texttt{Max} & 83.48\pmstd{0.30} & 89.04\pmstd{0.37} & 94.55\pmstd{0.09} & 89.88\pmstd{0.17} & 87.50\pmstd{0.26} & 90.87\pmstd{1.30} & 74.28\pmstd{1.27} & 87.09\pmstd{0.27}\\

\hline

\rule{0pt}{2.0ex}RoBERTa-base & \texttt{CLS} & 83.94\pmstd{0.30} & 90.44\pmstd{0.49} & 94.05\pmstd{0.06} & 90.70\pmstd{0.17} & 89.16\pmstd{0.22} & 90.80\pmstd{0.35} & 75.52\pmstd{0.42} & 87.80\pmstd{0.20}\\
\ & \texttt{Mean} & 84.88\pmstd{0.21} & 91.09\pmstd{0.01} & 94.60\pmstd{0.10} & 90.69\pmstd{0.07} & 89.73\pmstd{0.54} & 93.13\pmstd{0.12} & 77.22\pmstd{0.46} & 88.76\pmstd{0.08}\\
\ & \texttt{Max} & 83.98\pmstd{0.03} & 90.78\pmstd{0.24} & 93.96\pmstd{0.07} & 90.63\pmstd{0.11} & 90.05\pmstd{0.06} & 93.60\pmstd{0.72} & 77.80\pmstd{0.32} & 88.69\pmstd{0.12}\\

\hline

\rule{0pt}{2.0ex}RoBERTa-large & \texttt{CLS} & 85.63\pmstd{0.27} & 90.74\pmstd{0.15} & 94.53\pmstd{0.14} & 91.20\pmstd{0.11} & 90.08\pmstd{0.59} & 93.53\pmstd{0.76} & 72.66\pmstd{1.73} & 88.34\pmstd{0.28}\\
\ & \texttt{Mean} & 86.47\pmstd{0.29} & 91.53\pmstd{0.06} & 95.02\pmstd{0.08} & 91.15\pmstd{0.07} & 90.77\pmstd{0.34} & 92.33\pmstd{0.64} & 73.91\pmstd{0.96} & 88.74\pmstd{0.12}\\
\ & \texttt{Max} & 85.60\pmstd{0.26} & 90.73\pmstd{0.70} & 94.21\pmstd{0.65} & 91.09\pmstd{0.32} & 90.65\pmstd{0.37} & 91.53\pmstd{1.70} & 76.15\pmstd{0.33} & 88.56\pmstd{0.57}\\

\hline
\end{tabular}
\caption{
The percentage of correct answers (\%) for each task of SentEval.
The scores are the mean and standard deviation of three evaluations with different random seeds.
}
\label{table:senteval_full}
\end{table*}

\section{Average Runtime and Computing Infrastructure}
\label{appendix:runtime}
Fine-tuning for DefSent-BERT-base and DefSent-RoBERTa-base took about 5 minutes on a single NVIDIA GeForce GTX 1080 Ti.
Fine-tuning for DefSent-BERT-large and DefSent-RoBERTa-large took about 15 minutes on a single Quadro GV100.

\section{Full Results of the Word Prediction Experiment}
\label{appendix:d2w-full}
Table \ref{table:word_prediction_full} shows the experimental results on the word prediction experiment for each model and pooling strategy with learning rate.

\section{Predictions for definition sentences}
\label{appendix:d2w-actual}
Table \ref{tab:d2w_actual_results} shows the predicted words when the embeddings of definition sentences are input. We used BERT-large as a model and \texttt{CLS} as a pooling strategy for the experiment.
For prediction, sentences were first input into the model to obtain sentence embeddings. Then the sentence embeddings were input into the pre-trained word prediction layer to obtain word probabilities.
We show the top five words with the highest probability.

\section{Predictions for sentences other than definition sentences}
\label{appendix:d2w-arbitrary}
Table \ref{tab:d2w_arbitrary_results} shows the predicted words when the embeddings of sentences other than definition sentences are input.
We used BERT-large as a model and \texttt{CLS} as a pooling strategy for the experiment.
The evaluation procedure is the same as for Appendix \ref{appendix:d2w-actual}.

\section{Full Results of the STS Evaluation}
\label{appendix:sts-full}
Table \ref{table:sts_full} shows the experimental results on STS tasks for each model and pooling strategy.

\section{Full Results of the SentEval Evaluation}
\label{appendix:senteval-full}
Table \ref{table:senteval_full} shows the experimental results on SentEval tasks for each model and pooling strategy.

\end{document}